%% file: main.tex
\pdfoutput=1

\documentclass[11pt]{article}
\input{math_commands.tex}

\usepackage[final]{acl}

\usepackage{times}
\usepackage{latexsym}

\usepackage[T1]{fontenc}

\usepackage[utf8]{inputenc}

\usepackage{microtype}

\usepackage{inconsolata}

\usepackage{graphicx}

\definecolor{cadmiumgreen}{rgb}{0.0, 0.42, 0.24}
\definecolor{cardinal}{rgb}{0.77, 0.12, 0.23}
\definecolor{cadmiumred}{rgb}{0.89, 0.0, 0.13}

\usepackage{tcolorbox}
\newtcolorbox[list inside=prompt,auto counter,number within=section]{prompt}[1][]{
    fontupper=\ttfamily\footnotesize,
    boxsep=5pt,
    left=0pt,
    right=0pt,
    top=0pt,
    bottom=0pt,
    boxrule=1pt,
    #1,
}

\title{Mixed Signals: Decoding VLMs' Reasoning and Underlying Bias in Vision-Language Conflict}

\author{Pouya Pezeshkpour\\
Megagon Labs\\
\texttt{pouya@megagon.ai} \\
\And
Moin Aminnaseri\\
Megagon Labs\\
\texttt{moin@megagon.ai} \\
\And
Estevam Hruschka\\
Megagon Labs \\
\texttt{estevam@megagon.ai}}

\begin{document}
\maketitle
\begin{abstract}
Vision-language models (VLMs) have demonstrated impressive performance by effectively integrating visual and textual information to solve complex tasks. 
However, it is not clear how these models reason over the visual and textual data together, nor how the flow of information between modalities is structured.
In this paper, we examine how VLMs reason by analyzing their biases when confronted with scenarios that present conflicting image and text cues—a common occurrence in real-world applications.
To uncover the extent and nature of these biases, we build upon existing benchmarks to create five datasets containing mismatched image-text pairs, covering topics in mathematics, science, and visual descriptions. 
Our analysis shows that VLMs favor text in simpler queries but shift toward images as query complexity increases. This bias correlates with model scale, with the difference between the percentage of image- and text-preferred responses ranging from +56.8\% (image favored) to -74.4\% (text favored), depending on the task and model.
In addition, we explore three mitigation strategies: simple prompt modifications, modifications that explicitly instruct models on how to handle conflicting information (akin to chain-of-thought prompting), and a task decomposition strategy that analyzes each modality separately before combining their results. Our findings indicate that the effectiveness of these strategies in identifying and mitigating bias varies significantly and is closely linked to the model's overall performance on the task and the specific modality in question. We released our dataset and code\footnote{\url{https://github.com/megagonlabs/Modality-Bias}}
\end{abstract}

\section{Introduction}
Vision-language models (VLMs) have rapidly advanced the field of artificial intelligence by effectively combining visual and textual data to achieve state-of-the-art performance on a variety of tasks \citep{zhou2022learning, liu2023visual, zhang2024vision, li2025benchmark}. However, it remains unclear how these models integrate and reason over information from multiple modalities---a capability that becomes increasingly important in applications such as retrieval-augmented generation (RAG) \citep{yu2024visrag, yuan2024rag} and multi-agent systems \citep{ghafarollahi2024atomagents, jiang2024multi}, where data can be sourced from diverse modalities.

\begin{figure}[t!]
    \centering
    \includegraphics[width=\linewidth]{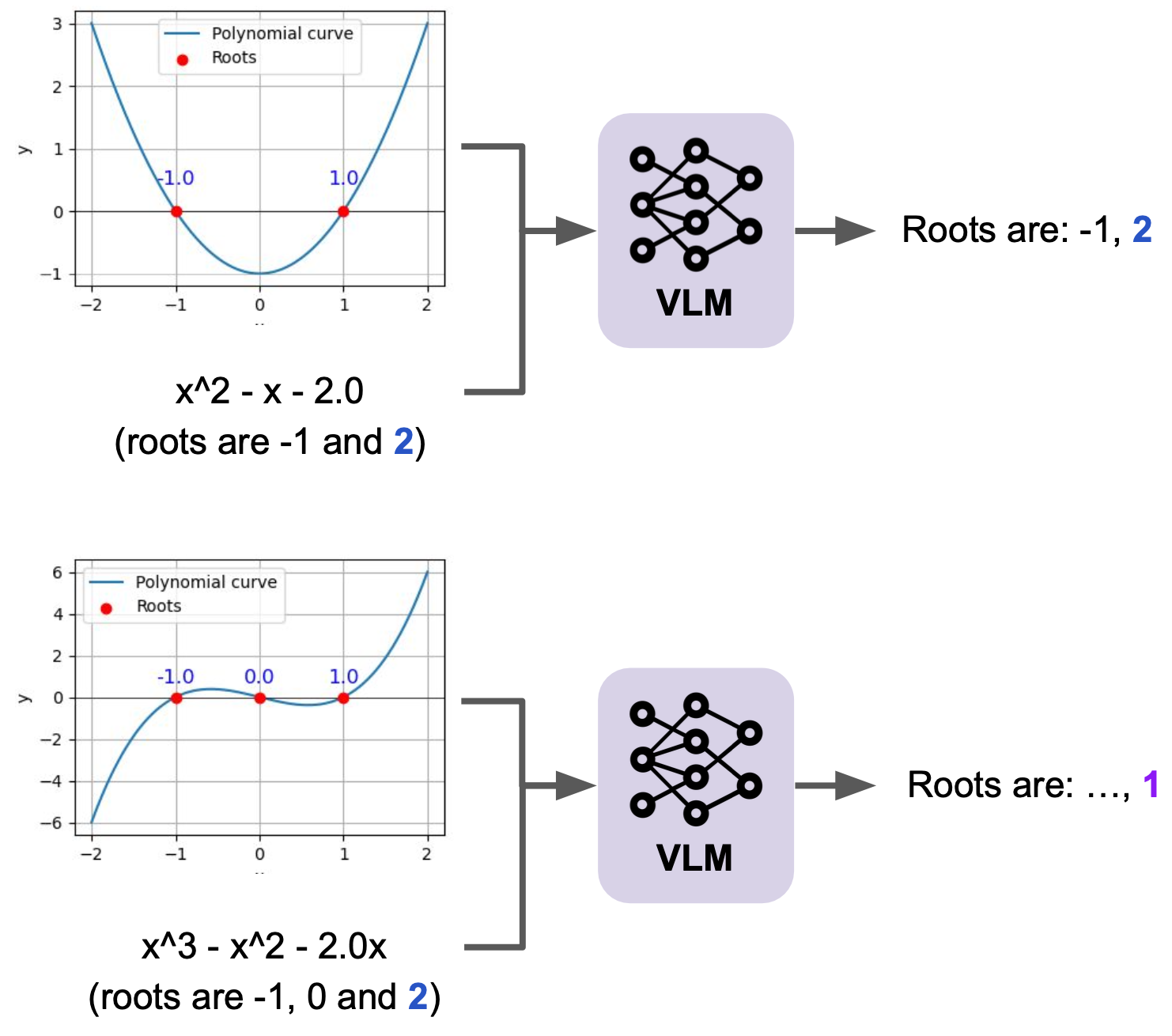}
    \caption{We investigate VLMs' bias toward text versus image inputs when mismatches occur between the modalities. Our observations reveal that this bias heavily depends on the task's difficulty. For example, while the model relies on textual representations to compute the roots of a degree-2 polynomial, increasing the degree to 3 shifts the reliance more toward the visual representation of the function.}
    \label{fig:illust}
    \postspace
\end{figure}
To understand how VLMs reason over multiple modalities, one approach is to investigate their behavior when confronted with conflicting information. Moreover, it is important to determine whether these models exhibit a bias toward one modality over the other, when exposed to deliberately mismatched image-text pairs (see Figure \ref{fig:illust}). For example, consider a scenario in healthcare where a chest X-ray image shows no signs of pneumonia, yet the accompanying text erroneously describes clear evidence of the disease. If the model disproportionately relies on the text, it may incorrectly diagnose pneumonia, potentially leading to inappropriate treatment recommendations and severe patient harm.

In this paper, we investigate VLMs' biases in the presence of conflicting multimodal cues by building upon existing datasets to create five novel benchmarks \citep{liu2023vsr,fu2024isobench}. Each benchmark consists of mismatched textual and image cues, with every sample assigned a complexity score to study the impact of sample difficulty on the models bias. 
These benchmarks span various topics, including mathematical reasoning, science, and visual descriptions. 
To generate the mismatched pairs, we begin with aligned image-text pairs from the original datasets and then employ a combination of rule-based methods and manual modifications to alter the textual content, thereby introducing conflicting information. 

We adopt five state-of-the-art VLMs spanning a diverse range of scales and conduct experiments on our constructed benchmarks. 
Our analysis reveals that when presented with conflicting information, VLMs tend to favor textual data over images in simpler scenarios. However, as query complexity increases, the models shift their bias toward the modality they perceive to be simpler---in many cases, this is the image. Furthermore, we observe that the extent of this bias correlates with the model's scale and strength. Finally, our findings indicate that this modality preference arises partly from the models’ perception of task difficulty and partly from other internal biases, which are strongly linked to their performance with different modalities as the input on a given task.

Observing the significant degree of modality bias in our experiments, we set out to explore several mitigation strategies aimed at reducing this issue. 
We consider three approaches to address modality bias: simple prompt modifications, introducing explicit instructions similar to chain-of-thought prompting, and a decomposition approach that analyzes each modality separately before combining outputs. Our experiments show that each approach exhibits a diverse impact based on the task, the model, and---most importantly---the model's performance when given different modalities as input. 
In scenarios where the model shows strong performance with both modalities (or even one modality) as the input, at least one of explored mitigation strategies demonstrate reasonable performance.
\section{Conflicting Modalities}
In this work, our goal is to understand how vision-language models (VLMs) reason over multiple modalities when they are presented with conflicting information. We simulate realistic scenarios in which the visual and textual inputs contradict one another, forcing the models to weigh and integrate disparate cues. In this section, we begin by outlining the problem setting, and then proceed to describe the methodology employed to construct each of our benchmarks. We provide the prompts used in creating the benchmark and the data statistics of created benchmarks in the Appendix.

\subsection{Problem Statement}
To evaluate how vision-language models (VLMs) handle conflicting multimodal cues, we formulate the following experimental setting. Assume we have a query \( Q \), an image \( I \), and a corresponding textual description \( T \) each yield a consistent answer \( A \). Each query is associated with a complexity score \( c \) that reflects the difficulty of the reasoning task. We then create a modified textual description \( T' \) that intentionally provides an answer \( A' \) different from \( A \).

In an ideal scenario, a VLM would detect the discrepancy between \( I \) and \( T' \) and indicate a conflict rather than committing to either \( A \) or \( A' \). However, if the model outputs an answer that aligns with one of the modalities (\( A \) or \( A' \)), this behavior is considered a bias toward that modality. We can quantify this bias by defining the bias metric \( B \) as:
\[
B = f(|A|,\, |A'|),
\]
where \(|\cdot|\) represents the number of times the model adheres to the response from a specific modality, and $f$ is a function measuring the difference between the inputs, which can be as simple as computing their ratio. This metric captures the degree to which the model’s output favors certain modality. In our experiments, we define $B$ as the difference between the percentage of image-favored responses and the percentage of text-favored responses.

\subsection{Benchmarking}
To systematically study VLM's biases, we build upon data from two existing datasets---VSR \citep{liu2023vsr} and Isobench \citep{fu2024isobench}---and create five distinct benchmarks featuring mismatched image-text pairs.

\paragraph{Graph Connectivity:}
For this benchmark, we start with graph samples sourced from Isobench. We manually modify the adjacency matrices in a minimal manner to alter the connectivity between target nodes. When the original graph is unconnected, we ensure that our modifications do not simply connect the two nodes directly, preserving the underlying structure while ensuring certain level of difficulty. We approximate the complexity of each sample in this task by the number of edges in its corresponding graph.

\paragraph{Function Convexity:}
Using samples from Isobench, we generate conflicting pairs by altering the functions expressions. Specifically, we multiply each coefficient of the function by minus one, creating a scenario where the textual description of the function’s convexity contradicts the visual representation. We approximate the complexity of each sample in this task based on the number of characters in the textual representation of the mathematical expression of the function.

\paragraph{Polynomial Roots Calculation:}
In this benchmark, we generate polynomials of degrees 1 through 4 (since they have closed form solution) by randomly selecting the roots within the range of –10 to 10 and use them to construct both the polynomial expressions and their corresponding visual representations. To create a conflicting pair, we randomly alter one of the roots by replacing it with a different value from the same range, resulting in a discrepancy between the textual and visual depictions of the polynomial. The degree of each polynomial serves as a proxy for task complexity.

\paragraph{Physics and Chemistry Questions:}
For this task, we leverage samples from Isobench and manually alter the textual descriptions of physics and chemistry problems. The modifications are designed so that the answer derived from the text would differ from that suggested by the visual cues, pointing to a choice other that the initial answer from the provided multiple-choices. Additionally, we filter out any questions that could be correctly answered by relying solely on the question, ensuring that the conflict between image and text is both meaningful and challenging. We manually assign an ``easy'' or ``hard'' label to each sample to reflect its complexity.

\paragraph{Visual Description:}
For this benchmark, we first select samples from the VSR dataset in which the original statement accurately reflects the content of the image. We then use GPT-4o mini to generate an extended description of the image centered around the given statement. By manually identifying the opposite of spatial relationships studied in the VSR dataset, we replace the original relation with its opposite (in the extended description), thereby creating a mismatched pair. We analyze the per-relation breakdown of VLMs performance as a proxy for task complexity.

\begin{figure}[t!]
    \centering
    \includegraphics[width=\linewidth]{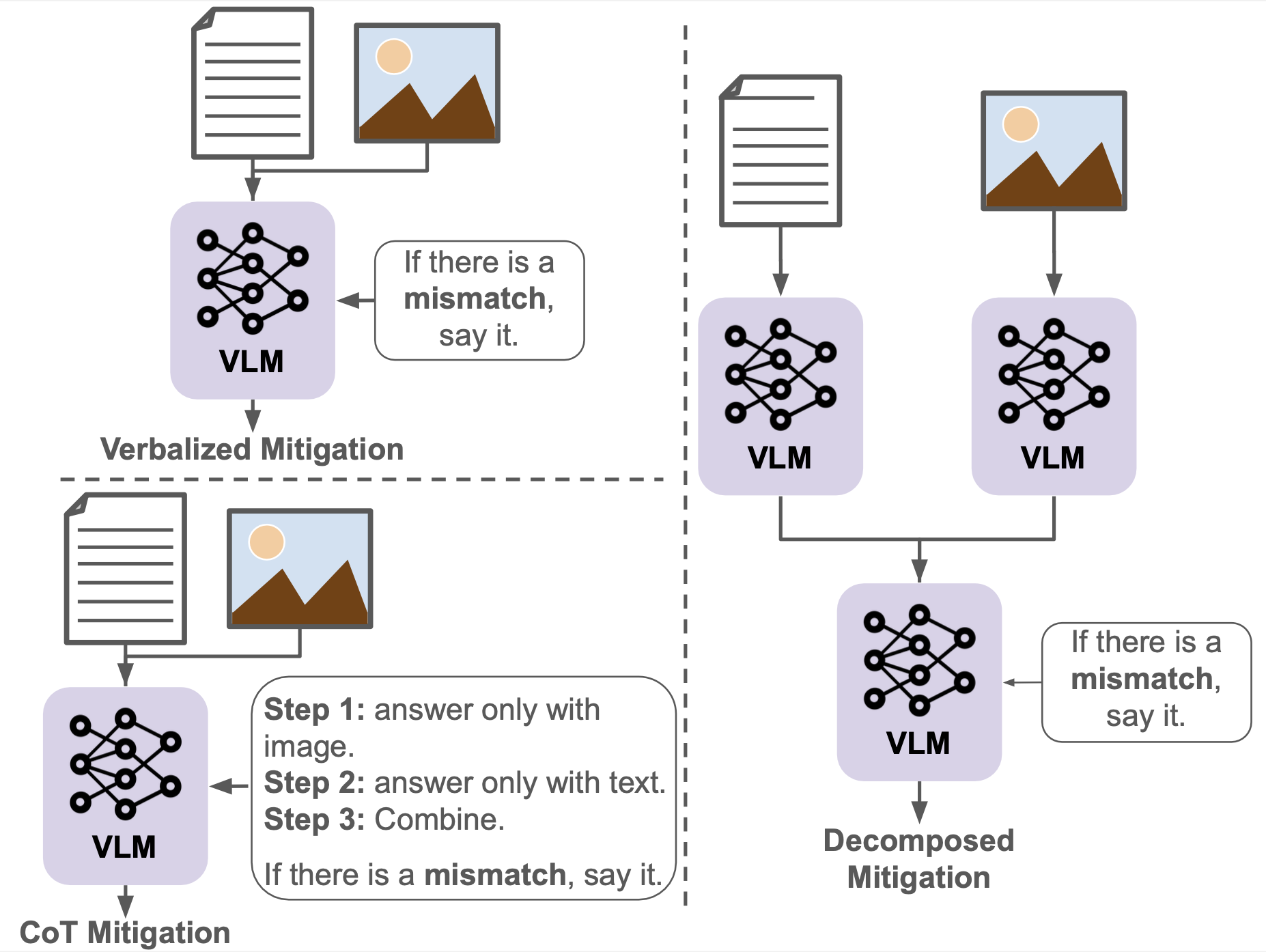}
    \caption{We investigate the impact of three mitigation strategies---Verbalized, CoT, and Decomposed---on identifying mismatches in the input modalities.}
    \label{fig:mitigation}
    \postspace
\end{figure}
\section{Mitigation Strategies}
Our experiments indicate that VLMs exhibit a notable bias when handling conflicting multimodal information (see section \ref{sec:bias}). To address this challenge, we propose three distinct mitigation strategies, as summarized in Figure \ref{fig:mitigation} with the adopted prompts provided in the Appendix. These strategies aim to detect and mitigate the bias by altering the models’ processing of visual and textual inputs, thereby encouraging a more balanced integration of information. The strategies are as follows:

\paragraph{Verbalized Mitigation:} Directly prompting the model to identify and report any mismatches or contradictions between the modalities. Instead of providing a conventional answer, the model is asked to simply indicate if it detects a discrepancy. This explicit acknowledgment of conflict helps prevent the model from defaulting to one modality.

\paragraph{CoT Mitigation:} Taking inspiration from chain-of-thought prompting, in this method, the model is guided through a three-step process. First, we instruct the model to process the image input alone; second, evaluate the textual description independently; and third, combine the outputs from both modalities. We further instruct the model to extract all relevant information from each modality, particularly when a single modality does not suffice to solve the task. Finally, the model is required to compare the two results and explicitly indicate if there is a mismatch.

\paragraph{Decomposed Mitigation:} In this method, we employ a multi-stage approach wherein the VLM is run three separate times. The first run solve the task using only the image input, the second only use the text input, and the third combines the outputs from the previous two runs. In the final stage, the model is specifically instructed to highlight any inconsistencies between the outputs, and if a discrepancy is detected, to report a mismatch.

\section{Experiments}
In here, we first examine the extent of bias in VLMs using our created benchmarks. Next, we examine how bias correlates with the models' perceived problem difficulty and conduct an error analysis of their performance. Finally, we explore the impact of various mitigation strategies in reducing the bias.

\subsection{Experimental Details}
For our experiments, we evaluate five state-of-the-art vision-language models: Qwen2-VL-7B-Instruct and Qwen2-VL-72B-Instruct \citep{wang2024qwen2}, Llam-3.2-90B-vision-instruct \citep{grattafiori2024llama}, GPT-4o mini, and GPT-4o \citep{hurst2024gpt}. 
These models have been selected for their diverse architectures and multimodal processing capabilities, enabling us to investigate how factors such as model scale, training methodology, and fusion strategies affect the integration of conflicting information. 
Moreover, we rely on accuracy and F1 scores as evaluation metrics. To determine the models' preferred modality for each sample, we identify the modality-specific label with which the output aligns. 
All prompts used with these models are provided in the Appendix.

\begin{figure*}[th!]
    \centering
    \begin{subfigure}[b]{0.7\textwidth}
        \centering
        \includegraphics[width=\linewidth]{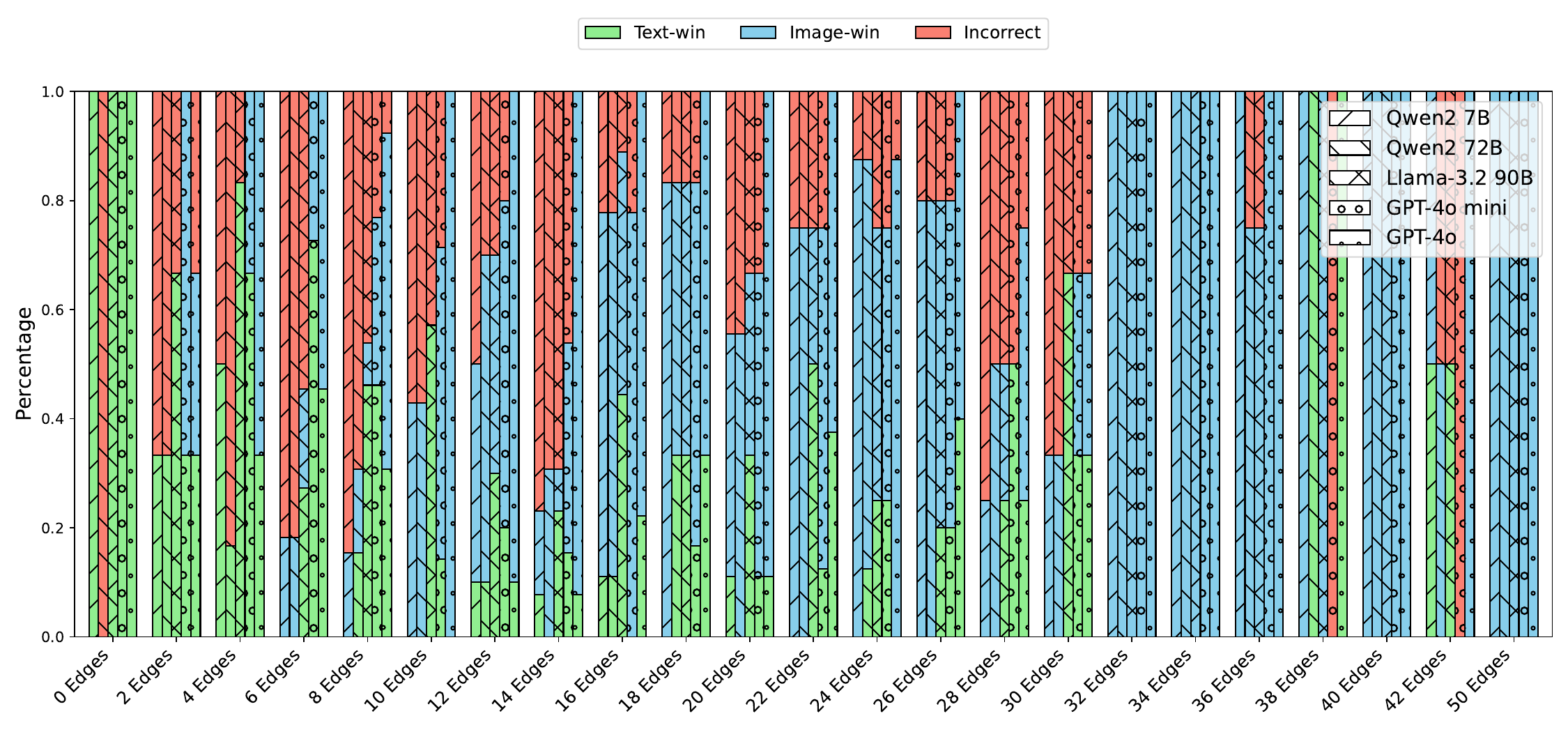}  
        \caption{Connectivity.}
        \label{fig:bias-connect}
    \end{subfigure}
    \begin{subfigure}[b]{0.28\textwidth}
        \centering
        \includegraphics[width=\linewidth]{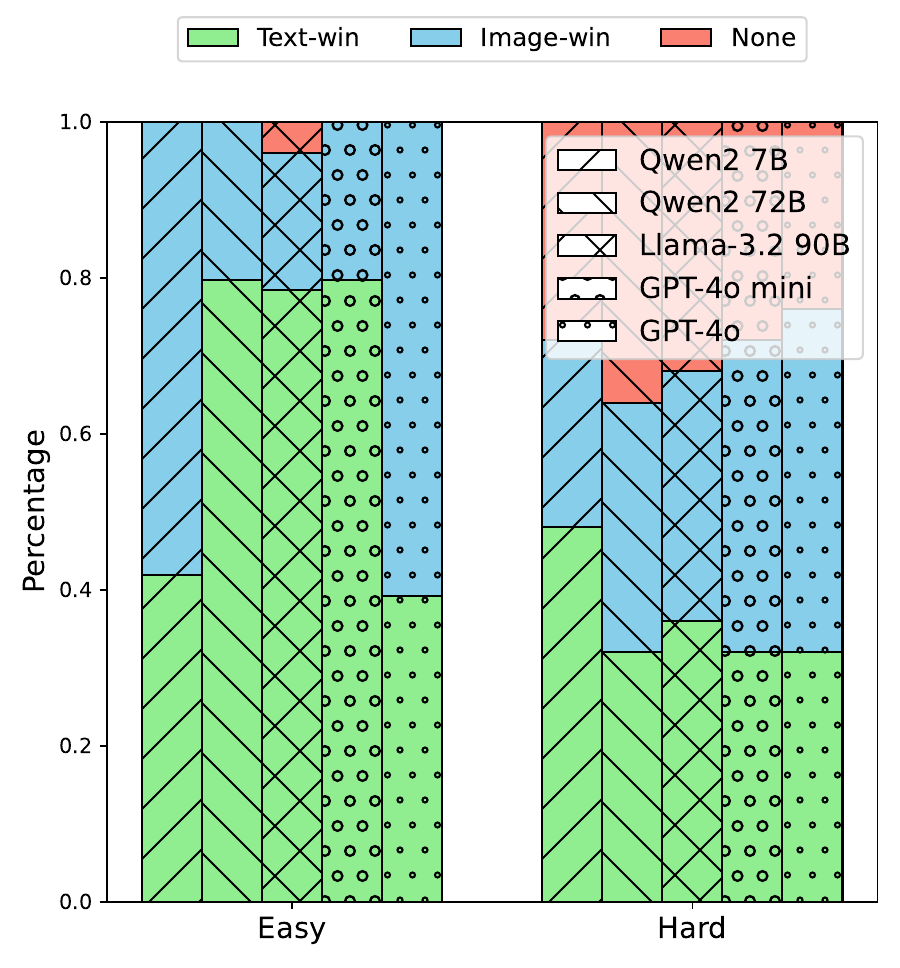}
        \vspace{-3pt}
        \caption{Science questions.}
        \label{fig:bias-science}
    \end{subfigure}
    \begin{subfigure}[b]{0.49\textwidth}
        \centering
        \includegraphics[width=\linewidth]{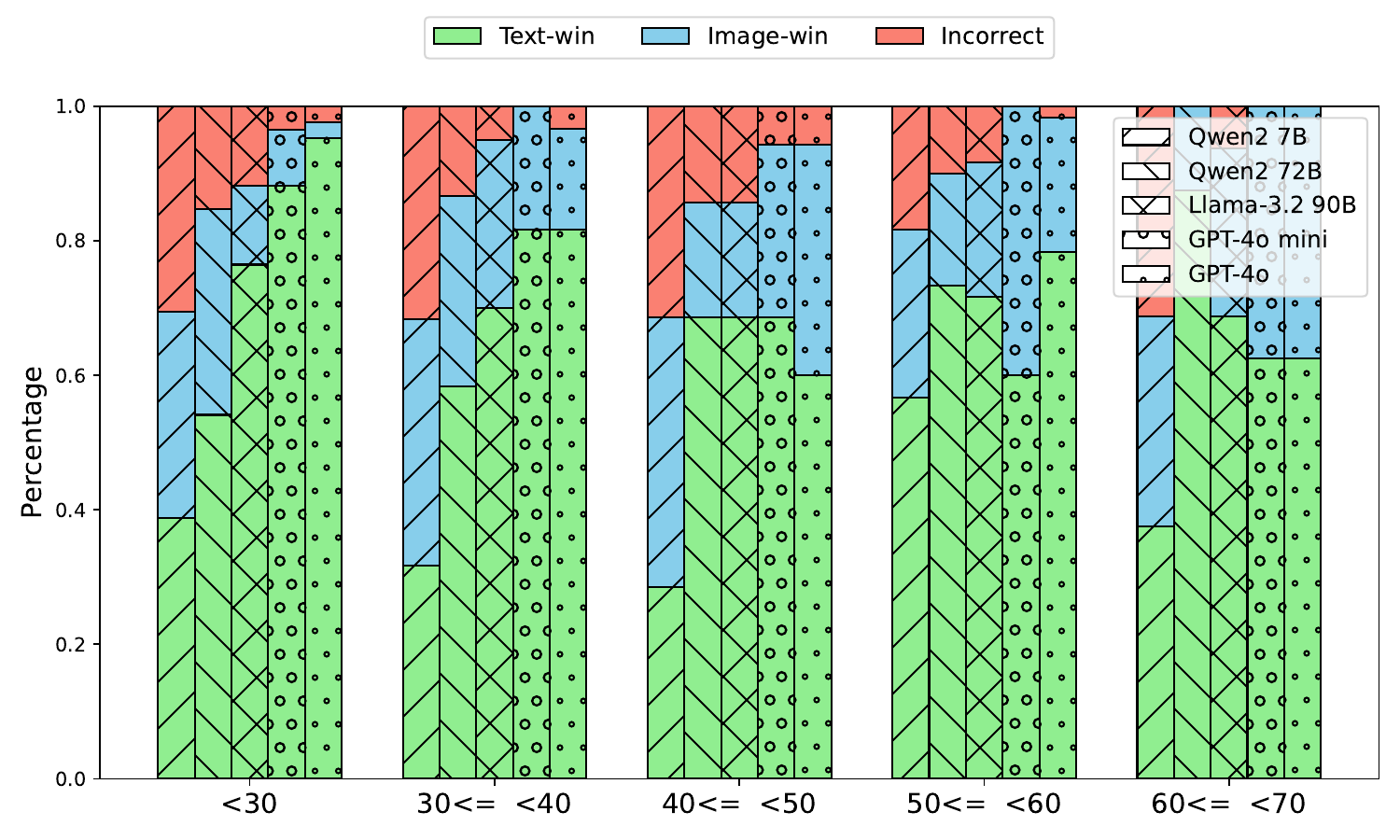}
        \caption{Convexity.}
        \label{fig:bias-convext}
    \end{subfigure}  
    \begin{subfigure}[b]{0.49\textwidth}
        \centering
        \includegraphics[width=\linewidth]{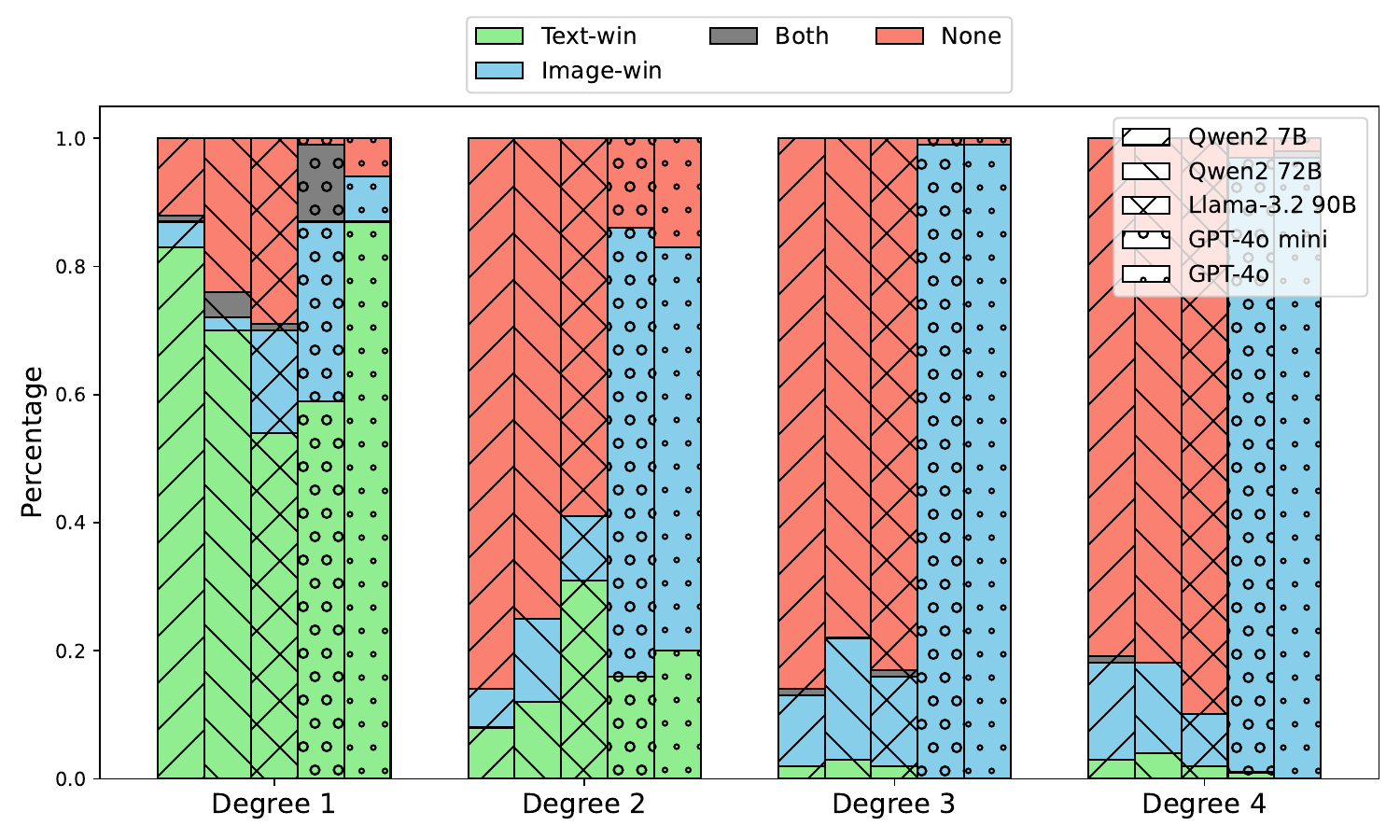}
        \caption{Roots calculation.}
        \label{fig:bias-root}
    \end{subfigure}     
    \caption{The distribution of VLMs biases toward text versus image inputs.}
    \label{fig:results}
    \minipostspace
\end{figure*}

\subsection{Bias in VLMs}
\label{sec:bias}
We measure the bias of VLMs toward text or image modalities using our created datasets. To isolate the impact of bias from the models mispredictions, we calculate the percentage of text versus image bias in mismatched inputs only for those samples where models produced correct predictions when both the image and text were aligned (i.e., the original inputs). Additionally, we report the percentage of incorrect predictions on the original, aligned inputs. We also present the accuracy of models' performance for solving each task when using only one modality or both, along with the corresponding $B$ value for each task and model in the Appendix.

\subsubsection{Bias in Mathematical Reasoning}

\paragraph{Graph Connectivity:} We present the distribution of modality preferences across graphs in Figure \ref{fig:bias-connect}, where the task is to determine whether two target nodes are connected. We categorize model performance by the number of edges in the graphs—a proxy for task difficulty. For simpler graphs with fewer edges, models tend to favor answers supported by textual input. However, as graph difficulty increases, there is a clear shift toward reliance on visual (image-based) information. Notably, for highly complex graphs (those with 32 edges or more), most models rely exclusively on image inputs to generate their responses. Moreover, while all models exhibit similar bias trends, Qwen models show a higher rate of incorrect predictions. Finally, based on the overall bias values reported in the Appendix (Table \ref{tab:b-val}), we observe that, with the exception of Llama-3.2 90B, all other models generally favor images over text in this task.

\paragraph{Function Convexity:} The bias distribution of VLMs in identifying the convexity of mathematical functions is depicted in Figure \ref{fig:bias-convext}. We approximate task difficulty based on the number of characters in each function's mathematical expression and divide the samples into five categories. The results indicate that for stronger models, as the difficulty increases, the textual bias decreases while the reliance on image cues increases. In contrast, for Qwen models, we observe a mixed impact of difficulty on modality bias. We suspect this inconsistency may be partly due to noise in our difficulty approximation method, as reflected by the higher number of incorrect predictions in some of the easier categories (those with fewer characters). 
Finally, based on the overall bias values reported in the Appendix (Table \ref{tab:b-val}), we find that all models generally favor text over image in this task.

\paragraph{Polynomial Roots Calculation:} The results of VLM bias in calculating the roots of polynomials are presented in Figure \ref{fig:bias-root}. As shown, there is a clear shift in modality preference as the degree of the polynomial increases. For polynomials of degree 1, the models exhibit a strong bias toward textual input, indicating that simpler algebraic reasoning is more effectively grounded in the accompanying text. In some instances, a few models even correctly identify the mismatch by reporting both the image- and text-derived roots. However, as we move to polynomials of degree 2 and higher, this reliance on textual information diminishes substantially, with models increasingly favoring visual (image-based) inputs. This transition likely reflects the growing complexity of the problem space, where interpreting graphical representations becomes more advantageous---or even necessary---for accurate problem solving. The models’ adaptive use of modalities suggests an emergent reasoning behavior, wherein they selectively leverage the most informative input source based on the task's complexity. 
Drawing on the overall bias values reported in the Appendix, we observe that OpenAI models tend to favor images, while other models generally favor text in this task.

\begin{figure*}[t!]
    \centering
    \includegraphics[width=\linewidth]{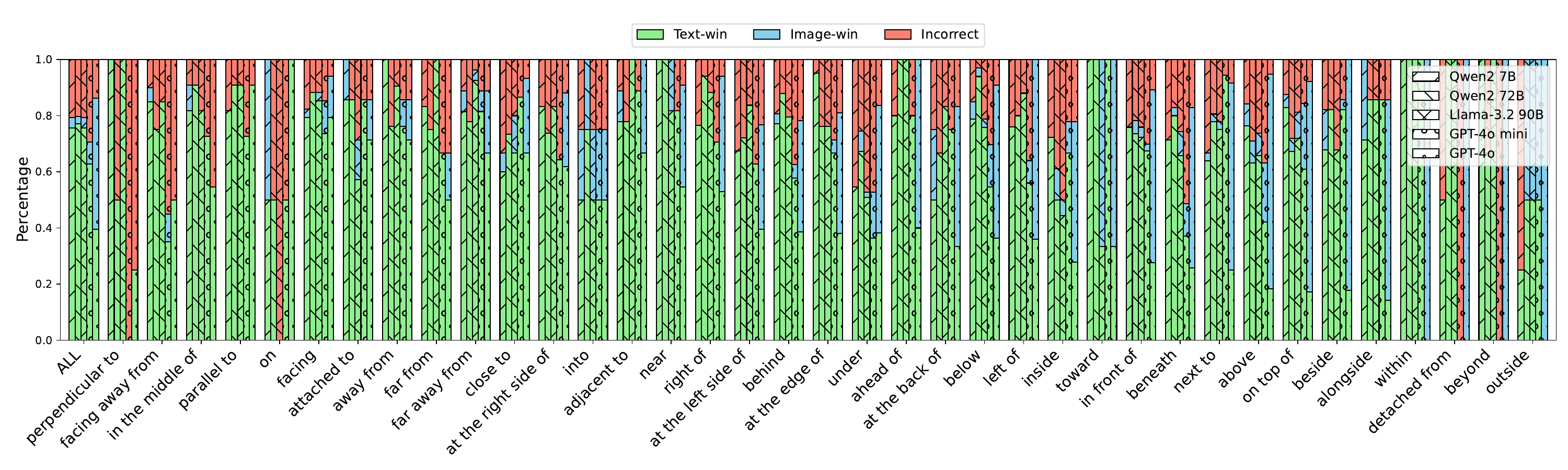}
    \caption{The distribution of VLM biases toward text versus image in the VSR-based dataset. We report the per- spatial relationship break down of performance.}
    \label{fig:bias-vsr}
    \minipostspace
\end{figure*}

\subsubsection{Bias in Science Questions} In answering scientific questions, we observe a notable reliance on textual information for easier questions, suggesting that the model is more confident in leveraging text when the reasoning demands are relatively low (see Figure \ref{fig:bias-science}). However, this reliance diminishes markedly for more challenging questions. Although support from image inputs appears to increase in these more challenging cases, a significant proportion of responses fall into the incorrect predictions. This trend suggests that, within the domain of science questions, the model struggles to produce a conclusive answer when faced with complex reasoning tasks that require effective integration of both text and image modalities. These findings potentially point to a limitation in the model’s ability to reconcile multimodal information under higher cognitive load. 
The overall trend in the bias values indicates that, except for Qwen2 7B and GPT-4o---which favor images---all other models favor text in this task (see Table \ref{tab:b-val}).

\subsubsection{Bias in Visual Description}
Investigating VLMs' biases toward text or image inputs using VSR samples, we observe distinct behaviors depending on the model and the specific spatial relation queried. For instance, GPT-4o exhibits a stronger bias toward textual data for spatial relations such as ``perpendicular to'', ``facing away from'',``in the middle of'', ``parallel to'', ``on'', ``facing'', and ``attached to''. In contrast, for relations like ``outside'', ``beyond'', ``detached from'', ``within'', ``alongside'', ``beside'', ``on top of'', and ``above'', the model shows a much greater bias toward image data, highlighting the strong impact of specific spatial relations on the VLMs modality preference. This variation is less pronounced in other models, although shifts in modality bias between text and image inputs are still evident across different spatial relations. We suspect that these differences largely reflect the training data, where the distribution of spatial relations varies across modalities. Based on the overall bias values reported in the Appendix (Table \ref{tab:b-val}), we observe that, except for GPT-4o, all other models generally favor text over image in this task.

\begin{figure}[t!]
    \centering
    \includegraphics[width=\linewidth]{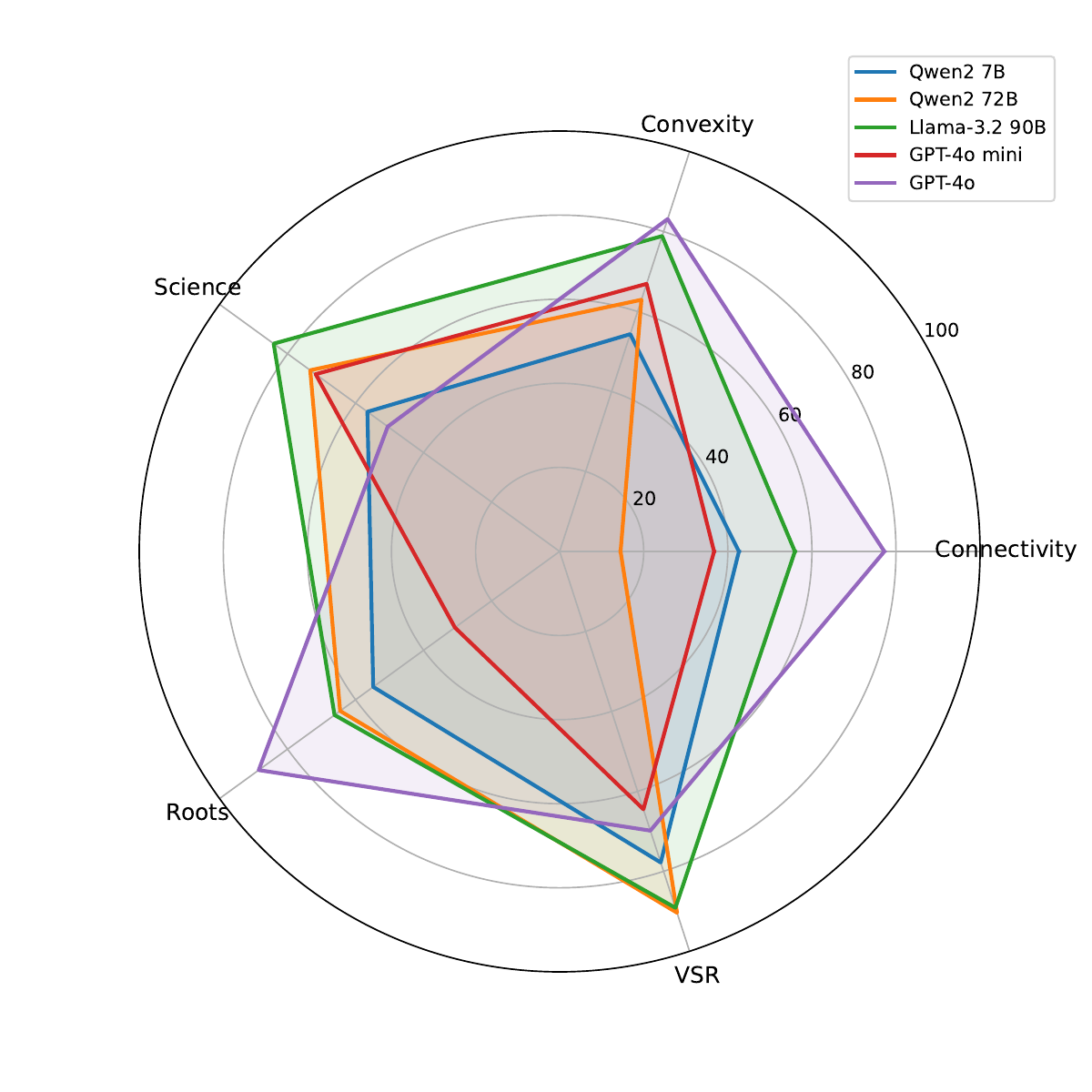}
    \vspace{-8mm}
    \caption{The accuracy of VLMs' internal perception of the simpler modality for solving the task is evaluated by comparing it to the actual modality each model relies on during problem solving.}
    \label{fig:difficulty}
    \postspace
\end{figure}
\subsection{VLMs Bias and Their Perceived Sample Difficulty}
In this section, we investigate the relationship between the models' inherent modality bias and their internal estimation of problem difficulty. Specifically, for each sample where the model provided a correct answer for initial matching inputs, we consider the modality it relied on (with mismatching inputs)---either text or image---as the gold label, representing the modality it implicitly deemed easier to use. 
Samples with incorrect initial answers are excluded to ensure that the gold label reflects a reliable internal assessment.
%
\begin{figure*}[t!]
    \centering
    \includegraphics[width=\linewidth]{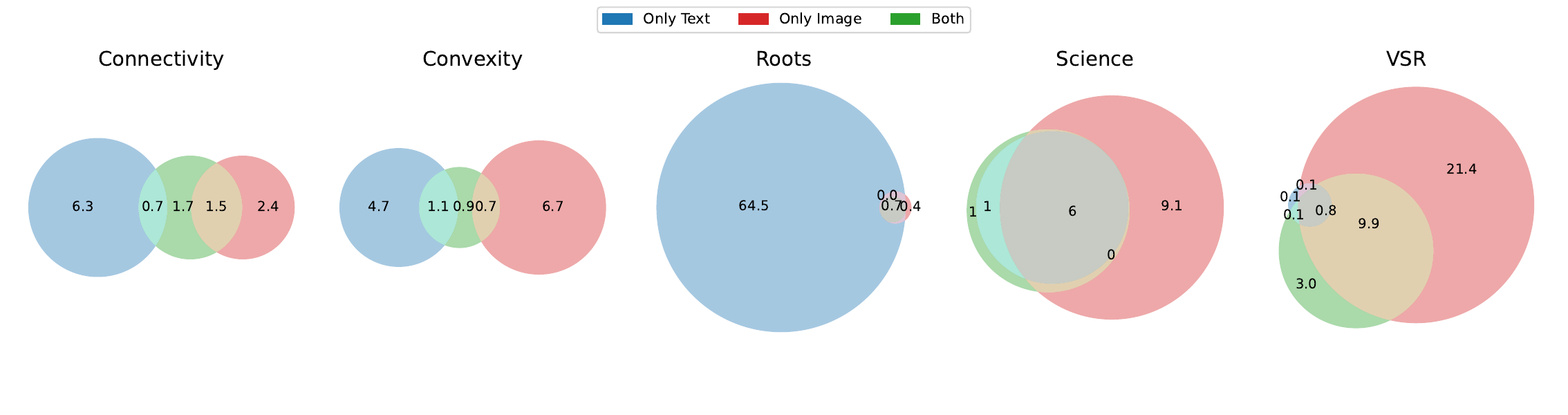}
    \vspace{-12mm}
    \caption{\textbf{Venn diagram of errors} displaying the percentage of mispredicted samples when GPT-4o is provided with only text, only image, and both text and image inputs.}
    \label{fig:gpt4o-error}
    \postspace
\end{figure*}
%
Subsequently, for each sample from our datasets, given the mismatched pair, we asked the models to explicitly predict which modality (text or image) they believed would be easier to use in answering the query. The accuracy of these predictions, which is summarized in Figure \ref{fig:difficulty}, reveals trends similar to those observed in our earlier analysis. In particular, stronger models show a clear alignment between their perceived ease 
of using a given modality and the bias they exhibit when resolving a task with conflicting visual and textual information. This correlation underscores the role of internal difficulty estimation in driving modality bias, suggesting that models tend to favor the modality they internally assess as less challenging for solving the task. 

In cases where we observe low accuracy---highly correlated with the models' overall task performance (see Table \ref{tab:acc} in the Appendix)---we suspect that additional internal biases influence the models' modality preferences. Analysis of the explanations provided by the models reveals that, in most of those instances, they do not rely on the specific details of the sample but rather use the general task description to determine which modality is easier to process.

\subsection{VLMs Bias and The Models Performance}
One possible approach to understanding the roots of modality bias in VLMs is through error analysis. By examining the correlation between model failures across different input modalities, we can gain deeper insights into the underlying mechanisms driving its decision-making. 
In Figure \ref{fig:gpt4o-error}, we present a Venn diagram illustrating GPT-4o’s error percentages under three different input settings: text-only, image-only, and combined matching initial text-image inputs (Venn diagrams of other models are provided in the Appendix). 
%
Analyzing the percentage of unique and overlapping errors (error independence) enables us to connect these failure modes to the biases observed earlier. For instance, if GPT-4o’s errors unique to text-only and image-only inputs are resolved when both modalities are provided, we can conclude that the model effectively integrates the two modalities and reason over them. On the other hand, errors unique to a single modality that persist even when both text and image inputs are provided reveal a blind bias toward that modality, further underscoring the model's inability to effectively reason across both modalities. Consequently, a higher proportion of such cases signals a stronger bias in favor of that modality.

As shown, the number of unique-modality errors that persist after combining both text and image inputs is higher for images in the connectivity and VSR tasks, while it is higher for text in the convexity task, highlighting greater model bias toward those modalities (this aligns with the overall bias values reported in Table \ref{tab:b-val} in the Appendix). Moreover, in the root calculation and science tasks, the number of persistent errors appears similar for both text-only and image-only inputs. 
Beyond highlighting these modality-specific failure patterns, diagrams also underscore how task complexity influences model performance. Specifically, GPT-4o's errors tend to cluster in a single modality for the root calculation, science, and VSR, whereas for connectivity and convexity tasks, errors are more evenly distributed between modalities. This suggests that the complexity of solving these tasks is more balanced between the modalities.

\begin{table*}[t!]
\small
\centering
\begin{tabular}{lrrrrrrrrrr}
\toprule 
\multirow{2}{*}{\bf Model} & \multicolumn{2}{c}{\bf Connectivity}&
\multicolumn{2}{c}{\bf Convexity}&\multicolumn{2}{c}{\bf Roots}&\multicolumn{2}{c}{\bf Science}&\multicolumn{2}{c}{\bf VSR}\\
\cmidrule(lr){2-3}
\cmidrule(lr){4-5}
\cmidrule(lr){6-7}
\cmidrule(lr){8-9}
\cmidrule(lr){10-11}
&ACC&F1&ACC&F1&ACC&F1&ACC&F1&ACC&F1\\
\midrule
Qwen2 7B + V& 49.6 & 7.1 & 53.5 & 36.7 & 50.3 & 14.6 & 58.5 & 51.7 & 57.0 & 68.0\\
Qwen2 7B + C& 50.0 & 0.0 & 49.8 & 16.2 & 49.8 & 8.2 & 50.5 & 15.5 & 64.6 & 70.5\\
Qwen2 7B + D&-&-&-&-&-&-&-&-&-&-\\
\midrule
Qwen2 72B + V& 52.7 & 48.9 & 50.1 & 3.7 & 52.1 & 44.2 & 61.6 & 40.6 & 76.0 & 79.0\\
Qwen2 72B + C& 48.8 & 25.9 & 50.9 & 27.2 & 52.7 & 33.6 & 54.5 & 19.6 & 80.5 & 79.0\\
Qwen2 72B + D & 49.6 & 25.4 & 54.8 & 62.0 & 50.6 & 64.9 & 62.6 & 63.7 & 23.2 & 24.7\\
\midrule
Llama-3.2 90B + V& 50.7 & 41.6 & 56.4 & 60.1 & 55.3 & 62.8 & 68.1 & 58.2 & 45.8 & 35.1\\
Llama-3.2 90B + C& 51.1 & 12.5 & 54.1 & 32.2 & 52.6 & 44.6 & 57.0 & 30.8 & 46.6 & 19.3\\
Llama-3.2 90B + D & 52.7 & 52.5 & 60.7 & 62.9 & 53.5 & 63.6 & 68.1 & 71.2 & 10.2 & 9.9\\
\midrule
GPT-4o mini + V&51.5&17.3&70.9&63.7&71.4&65.8&63.1&45.1&74.6&\bf 79.2\\
GPT-4o mini + C&56.3&39.8&82.8&82.2&\bf 72.1&63.1&61.6&40.6&54.6&55.6\\
GPT-4o mini + D&73.8&73.1&83.6&84.4&70.4&\bf 77.1&69.2&70.5&41.6&45.7\\
\midrule
GPT-4o + V&67.6&55.6&\bf 87.3&\bf 86.8&72.0&62.7&78.3&75.7&\bf 76.3&77.2\\
GPT-4o + C&58.6&30.3&82.8&80.4&69.3&56.7&75.3&68.4&73.9&72.2\\
GPT-4o + D&\bf 81.3&\bf 80.8&85.0&85.9&68.3&75.9&\bf 79.3& \bf 79.4&68.7&68.7\\
\bottomrule
\end{tabular}
\caption{\textbf{The performance of the mitigation strategies} is evaluated by calculating the accuracy and F1 metrics for the three approaches, Verbalized (V), CoT (C), and Decomposed (D). We assess their effectiveness in reporting ``mismatch'' when a discrepancy exists between the input modalities.}
\label{tab:mitigate}
\postspace
\end{table*}
\subsection{Mitigating Bias in VLMs}
To investigate the impact of our mitigation strategies, we calculate their accuracy in identifying mismatches over combination of initially matched and modified mismatched text-image pairs for each dataset (resulting in balanced evaluation sets). 
The results of bias mitigation strategies are summarized in Table \ref{tab:mitigate} (the Qwen2 7B model was incapable of solving the task when the image was missing). Our findings demonstrate that if models achieve high accuracy on the original task, at least one of the three proposed strategies—Verbalized, CoT, and Decomposed—can effectively detect mismatches between visual and textual inputs. Notably, stronger models like GPT-4o achieve particularly high accuracy in conflict detection. 

Our analysis indicates that the effectiveness of each mitigation strategy varies by both model and task. For instance, the decomposed method only appears effective when the model's performances using image-only and text-only inputs are reasonably high. Moreover, although both the Verbalized and CoT approaches excel when the model's performance on the original task is high, they exhibit distinct behavior: the Verbalized approach performs better for open models (despite overall low accuracy) and GPT-4o, whereas GPT-4o mini benefits more from the CoT method in the connectivity, convexity, and root calculation tasks. 
These differences highlight the nuanced nature of bias in VLM predictions and suggest that tailored mitigation techniques may be necessary to address specific challenges. Overall, our results underscore the need for a multifaceted approach to mitigate modality bias, ultimately enhancing the reliability and fairness of vision-language models.

\section{Related Works}
As LLMs and VLMs become more capable in reasoning tasks \citep{zhang2023multimodal,ahn2024large,davoodi2024llms,wang2024picture}, it is essential to investigate the interplay between modalities and how models reason across them to solve tasks. 
Previous studies have examined various aspects of VLM reasoning capabilities by introducing benchmarks that assess their general performance \citep{liu2024mmbench,yue2024mmmu}, or by focusing on specific abilities such as spatial reasoning \citep{chen2024spatialvlm} and robot navigation \citep{zeng2023large}. Despite these efforts, it remains unclear how these models reason and combine information across multiple modalities, and whether they exhibit specific biases toward any input modality.

Many previous works investigate the robustness of VLMs \citep{chang2024survey}. For instance, \citet{yuksekgonul2022and} demonstrate that VLMs are sensitive to the order of words, while \citet{dumpala2024sugarcrepe++} show that they struggle to distinguish between semantically equivalent but lexically different captions. Furthermore, several studies have examined VLMs robustness through adversarial attacks \citep{liu2024survey,ye2025survey} and jailbreak strategies \citep{tao2024imgtrojan, jin2024jailbreakzoo}. 
The study most closely related to ours is a concurrent work \citep{deng2025words} that examines VLMs bias toward text and image inputs in conflicting scenarios. Our approach differs in three key aspects. First, the authors in \citep{deng2025words} generate mismatching text using GPT-4o; in contrast, we create mismatched text using rule-based and manual methods, giving us greater control. Second, while \citep{deng2025words} report that VLMs exhibit a ``blind faith in text'', we demonstrate that VLMs can, in fact, show a stronger bias toward images in certain tasks, and the extent of their bias is heavily correlated with sample difficulty. Third, whereas their work focuses on supervised fine-tuning approaches for mitigation, which can be challenging and expensive, we explore the impact of post-processing mitigation strategies. 

\section{Conclusion}
We investigate bias in VLMs by creating five novel benchmarks that span various tasks and domains, including mathematical reasoning, science, and visual description. Our findings reveal that VLMs exhibit distinct biases toward either textual or visual cues depending on the task, with the nature of this bias shifting according to sample difficulty and model capabilities---favoring text in simpler scenarios and images in more complex ones. Furthermore, we explore a range of mitigation strategies—including verbalized, chain-of-thought, and decomposed approaches—which have demonstrated reasonable accuracy in detecting mismatches, provided the model achieves strong performance on the original task. These results underscore the complexity of multimodal reasoning in VLMs and highlight promising directions for enhancing their reliability in real-world applications.

\section{Limitations}
Despite the insights provided by our study, several limitations must be acknowledged. First, our analysis is restricted to only five vision-language models. While these models represent a range of architectures and capabilities, they may not capture the full diversity of VLM behaviors present in the broader research community. Future work should expand this analysis to include additional models to validate and generalize our findings.

In addition, although our five benchmarks cover various tasks and domains---including mathematical reasoning, science, and visual descriptions---they may not fully represent the myriad of real-world scenarios where multimodal conflicts occur. Our benchmarks, while comprehensive in their scope, are still limited in terms of the diversity and complexity of tasks that can arise in practical applications. 

Finally, while the mitigation strategies we investigated (Verbalized, CoT, and Decomposed) have shown promising results in some cases, they may not be universally applicable across all models and tasks. Further research is needed to refine these approaches and explore additional strategies that could enhance the robustness and fairness of vision-language models in handling conflicting information.

\bibliography{custom}

\appendix

\section{Data}
We build our dataset on top of two existing sources. The first, VSR, contains over 10k natural text (one-sentence) and image pairs, annotated with 66 types of spatial relations and true/false labels indicating whether the text and image are aligned. The second, Isobench, is a dataset featuring problems from math, science, algorithms, and games, with each example presented in multiple isomorphic formats (e.g., visual and textual). The data statistics for our created datasets are provided in Table \ref{tab:stat}.

\section{Prompts}
The prompt used to extend the VSR description is provided in \ref{prompt:vsr}. Additionally, the prompt for solving tasks using both image and text inputs is presented in \ref{prompt:conn-pair}, \ref{prompt:conv-pair}, \ref{prompt:root-pair}, \ref{prompt:science-pair} and \ref{prompt:vsr-pair}. We also include the prompts for solving the graph connectivity task with single modality inputs in \ref{prompt:conn-t} and \ref{prompt:conn-i}. Moreover, the prompts employed in the mitigation strategies for graph connectivity are provided in \ref{prompt:mit-v}, \ref{prompt:mit-c}, \ref{prompt:mit-d-c}, \ref{prompt:mit-d-i}, and \ref{prompt:mit-d-t}.

\begin{table}[t!]
\small
\centering
\begin{tabular}{lr}
\toprule 
\bf Task & \bf \# Samples\\
\midrule
Connectivity& 128\\
Convexity& 256\\
Roots& 400\\
Science& 99\\
VSR& 846\\
\bottomrule
\end{tabular}
\caption{Data statistics of the created benchmarks.}
\label{tab:stat}
\postspace
\end{table}
\begin{prompt}[title={\footnotesize\texttt{VSR Description Extension}}, label=prompt:vsr]
Below is a sentence that serves as the central idea, along with a specified relationship between entities within that sentence. Given an image, your task is to create a complete, coherent paragraph by adding few sentences before and after the given sentence describing the image. The additional sentences should establish context in a logical manner. However, these extra sentences should not directly connect to the provided relationship. Ensure that the specific relationship appears exactly once as given within the paragraph. Finally, do not mention `image' in the paragraph. \\
\\
Given sentence:\\
\{\}
\\
Specified relationship:\\
\{\}
\end{prompt}

\begin{prompt}[title={\footnotesize\texttt{Connectivity - Pair Input}}, label=prompt:conn-pair]
You are given the adjacency matrix of a graph and two target nodes. The goal is to determine whether the target nodes are connected. Additionally, you are provided with an image of the graph, where the target nodes are highlighted in \{\}. Use both the image and the adjacency matrix to answer the question.\\ 
The output should be in the following format:\\
Explanation: <....>\\
Answer: <True or False>\\
\\
Adjacency matrix: \\
\{\}\\
\\
Target nodes = \{\} and \{\}
\end{prompt}

\begin{prompt}[title={\footnotesize\texttt{Convexity - Pair Input}}, label=prompt:conv-pair]
You are given the mathematical expression of a function. The goal is to determine whether the function is convex or concave. Additionally, you are provided with the plot of the function. Use both the plot and the mathematical expression of the function to identify the convexity of the function.\\
The output should be in the following format:\\
Explanation: <....>\\
Answer: <convex or concave>\\
\\
The mathematical expression: \\
\{\}
\end{prompt}

\begin{prompt}[title={\footnotesize\texttt{Roots - Pair Input}}, label=prompt:root-pair]
You are given the mathematical expression of a polynomial. The goal is to find the roots of this polynomial. Additionally, you are provided with an image of the plot of that polynomial, where the roots are highlighted in red. Use both the image and the the mathematical expression of a polynomial to find the roots. \\
The output should be in the following format:\\
Explanation: <....>\\
Answer: <[The list of roots, comma separated]>\\
\\
The polynomial: \\
\{\}
\end{prompt}

\begin{prompt}[title={\footnotesize\texttt{Science - Pair Input}}, label=prompt:science-pair]
You are given a question, multiple choices, a textual description, and an image. Your task is to analyze both the text and the image and then choose the answer of the question from the provided choices.\\
The output should be in the following format:\\
Explanation: <....>\\
Answer: <A or B or ...>\\
\\
The textual description: \\
\{\}\\
\\
The question:\\
\{\}\\
\\
The Choices:\\
\{\}
\end{prompt}

\begin{prompt}[title={\footnotesize\texttt{VSR - Pair Input}}, label=prompt:vsr-pair]
You are given a statement, a paragraph, and an image. Your task is to analyze both the paragraph and the image to decide whether the statement is true or false.\\
The output should be in the following format:\\
Explanation: <....>\\
Answer: <true or false>\\
\\
The paragraph: \\
\{\}\\
\\
The statement:\\
\{\}
\end{prompt}

\begin{prompt}[title={\footnotesize\texttt{Connectivity - Only Text}}, label=prompt:conn-t]
You are given the adjacency matrix of a graph and two target nodes. The goal is to determine whether the target nodes are connected. \\ 
The output should be in the following format:\\
Explanation: <....>\\
Answer: <True or False>\\
\\
Adjacency matrix: \\
\{\}\\
\\
Target nodes = \{\} and \{\}
\end{prompt}
\begin{prompt}[title={\footnotesize\texttt{Connectivity - Only Image}}, label=prompt:conn-i]
You are provided with an image of a graph, with two target nodes highlighted in \{\}. The goal is to determine whether the target nodes are connected. \\ 
The output should be in the following format:\\
Explanation: <....>\\
Answer: <True or False>\\
\\
Target nodes = \{\} and \{\}
\end{prompt}

\begin{prompt}[title={\footnotesize\texttt{Mitigation-V - Connectivity}}, label=prompt:mit-v]
You are given the adjacency matrix of a graph and two target nodes. The goal is to determine whether the target nodes are connected. Additionally, you are provided with an image of the graph, where the target nodes are highlighted in \{\}. Use both the image and the adjacency matrix to answer the question. \\
If there is a mismatch or a contradiction between image and text, the answer should be `Mismatch'. \\
The output should be in the following format:\\
Explanation: <....>\\
Answer: <True or False or Mismatch>\\
\\
Adjacency matrix: \\
\{\}
\\
Target nodes = \{\} and \{\}
\end{prompt}

\begin{prompt}[title={\footnotesize\texttt{Mitigation-C - Connectivity}}, label=prompt:mit-c]
You are given the adjacency matrix of a graph and two target nodes. The goal is to determine whether the target nodes are connected. Additionally, you are provided with an image of the graph, where the target nodes are highlighted in \{\}. Use both the image and the adjacency matrix to answer the question. \\
\\
Follow these steps to find the answer:\\
\\
Step 1: Attempt to determine the answer using only the image. If the image alone does not provide the answer, extract all relevant information from it.\\
Step 2: Attempt to determine the answer using only the text. If the text alone does not provide the answer, extract all relevant information from it.\\
Step 3: Compare the answers or extracted information from both the text and the image, combine them, and derive the final answer. \\
\\
If there is a mismatch or a contradiction between image and text, the answer should be `Mismatch'. \\
The output should be in the following format:\\
Explanation: <....>\\
Answer: <True or False or Mismatch>\\
\\
Adjacency matrix: \\
\{\}
\\
Target nodes = \{\} and \{\}
\end{prompt}

\begin{prompt}[title={\footnotesize\texttt{Mitigation-D (Combining Module) - Connectivity}}, label=prompt:mit-d-c]
Given the adjacency matrix of a graph, two target nodes, and the image of the graph, the goal was to determine whether the target nodes are connected.\\
We provide only the task's textual description to an LLM to obtain an answer, and separately supply only the visual description to a VLM to obtain its answer.\\
Your goal is to compare the answers from the both models, combine them, and derive the final answer.\\ 
If there is a mismatch or a contradiction between two answers, the final answer should be `Mismatch'.\\
The output should be in the following format:\\
Explanation: <....>\\
Answer: <True or False or Mismatch>\\
\\
LLM's Answer: \\
\{\}\\
\\
VLM's Answer: \\
\{\}
\end{prompt}

\begin{prompt}[title={\footnotesize\texttt{Mitigation-D (Image Module) - Connectivity}}, label=prompt:mit-d-i]
You are provided with an image of a graph, with two target nodes highlighted in \{\}. The goal is to determine whether the target nodes are connected.\\
If the provided information is insufficient to solve the task, extract all relevant details from the input.\\
The output should be in the following format:\\
Explanation: <....>\\
Answer: <True or False>\\
\\
Target nodes = \{\} and \{\}\\
\end{prompt}

\begin{prompt}[title={\footnotesize\texttt{Mitigation-D (Text Module) - Connectivity}}, label=prompt:mit-d-t]
You are given the adjacency matrix of a graph and two target nodes. The goal is to determine whether the target nodes are connected.\\
If the provided information is insufficient to solve the task, extract all relevant details from the input.\\
The output should be in the following format:\\
Explanation: <....>\\
Answer: <True or False>\\
\\
Adjacency matrix: \\
\{\}\\
\\
Target nodes = \{\} and \{\}\\
\end{prompt}

\section{VLMs Performance on The datasets}
We provide the accuracy of VLMs on each task in Table \ref{tab:acc}. For the root calculation task, we report the average accuracy. First, the performance on both the original and modified text is nearly identical across all models, highlighting the minimal modifications applied to create conflicting textual cues. In most cases, models perform better using textual representations alone compared to using images alone, which aligns with observations from the original papers. However, for connectivity, GPT-4o, for convexity, Qwen2 72B, and for root calculation, both GPT-4o and GPT-4o mini, the image input appears to be more helpful than text alone. Finally, Qwen2 7B was unable to solve the task when the image was missing.

\begin{table*}[t!]
\small
\centering
\begin{tabular}{llrrrr}
\toprule 
&Model & Only Text Original& Only Text Modified&Only Image&Pair Text-Image Original\\
\midrule
\multirow{5}{*}{\rotatebox[origin=c]{90}{\bf Connectivity}}&Qwen2 7B&- & - & 48.4 & 53.1\\
&Qwen2 72B& 50.7 & 54.6 & 50.7 & 53.9\\
&Llama-3.2 90B&57.8 & 58.5 & 55.4 & 65.6\\
&GPT-4o mini&86.7 & 80.4 & 84.3 & 76.5\\
&GPT-4o&92.9 & 84.3 & 96.0 & 96.0\\
\midrule
\multirow{5}{*}{\rotatebox[origin=c]{90}{\bf Convexity}}&Qwen2 7B& - & - & 19.5 & 71.8\\
&Qwen2 72B&61.7 & 60.5 & 73.4 & 87.5\\
&Llama-3.2 90B&85.9 & 86.3 & 54.2 & 90.6\\
&GPT-4o mini&95.3 & 95.3 & 91.4 & 98.0\\
&GPT-4o&94.1 & 90.2 & 92.5 & 97.2\\
\midrule
\multirow{5}{*}{\rotatebox[origin=c]{90}{\bf Roots}}&Qwen2 7B& - & - & 31.2 & 36.1\\
&Qwen2 72B&28.2 & 29.6 & 24.6 & 33.8\\
&Llama-3.2 90B&31.6 & 29.3 & 16.7 & 35.6\\
&GPT-4o mini&38.6 & 37.4 & 96.8 & 97.5\\
&GPT-4o&34.7 & 34.3 & 98.8 & 99.2\\
\midrule
\multirow{5}{*}{\rotatebox[origin=c]{90}{\bf Science}}&Qwen2 7B& - & - & 60.6 & 84.8\\
&Qwen2 72B&89.8 & 86.8 & 54.5 & 89.8\\
&Llama-3.2 90B&87.8 & 84.8 & 50.5 & 89.8\\
&GPT-4o mini&87.8 & 82.8 & 67.6 & 87.8\\
&GPT-4o&92.9 & 89.8 & 84.8 & 91.9\\
\midrule
\multirow{5}{*}{\rotatebox[origin=c]{90}{\bf VSR}}&Qwen2 7B& - & - & 7.8 & 79.3\\
&Qwen2 72B&97.5 & 97.9 & 30.6 & 79.6\\
&Llama-3.2 90B&93.1 & 98.3 & 12.6 & 79.3\\
&GPT-4o mini&96.6 & 99.1 & 35.9 & 70.5\\
&GPT-4o&98.8 & 95.9 & 67.7 & 86.1\\
\bottomrule
\end{tabular}
\caption{The accuracy of models in solving each task when provided with text and image inputs independently and in combination.}
\label{tab:acc}
\end{table*}

\section{VLMs Bias Value}
We calculate the bias value for VLMs as the difference between the percentage of image-favored responses and the percentage of text-favored responses, with results provided in Table \ref{tab:b-val}. Positive values indicate a preference for images, while negative values reflect a preference for textual input. Our findings reveal significant variation across models and tasks, with bias values ranging from 56.8\% in favor of images to 74.4\% in favor of text. In general, less capable models tend to exhibit a stronger bias toward textual data.  Moreover, VLMs show the highest bias toward text in the VSR dataset, whereas they exhibit a greater degree of bias toward images in the graph connectivity task.

\begin{table*}[t!]
\small
\centering
\begin{tabular}{lrrrrr}
\toprule 
Model & Connectivity& Convexity&Roots&Science&VSR\\
\midrule
Qwen2 7B&37.5&-7.8&-44.4&6.0&-71.9\\
Qwen2 72B&38.2&-39.8&-29.0&-44.4&-74.4\\
Llama-3.2 90B&-7.8&-53.9&-29.4&-46.4&-71.9\\
GPT-4o mini&23.4&-53.5&56.8&-42.4&-54.9\\
GPT-4o&52.3&-65.2&42.3&19.1&6.9\\
\bottomrule
\end{tabular}
\caption{\textbf{VLMs' bias value} for each dataset, calculated as the difference between the percentage of image-favored responses and the percentage of text-favored responses. Positive values indicate an image preference, while negative values indicate a text preference.}
\label{tab:b-val}
\end{table*}

\section{Error Analysis}
\begin{figure*}[t!]
    \centering
    \includegraphics[width=\linewidth]{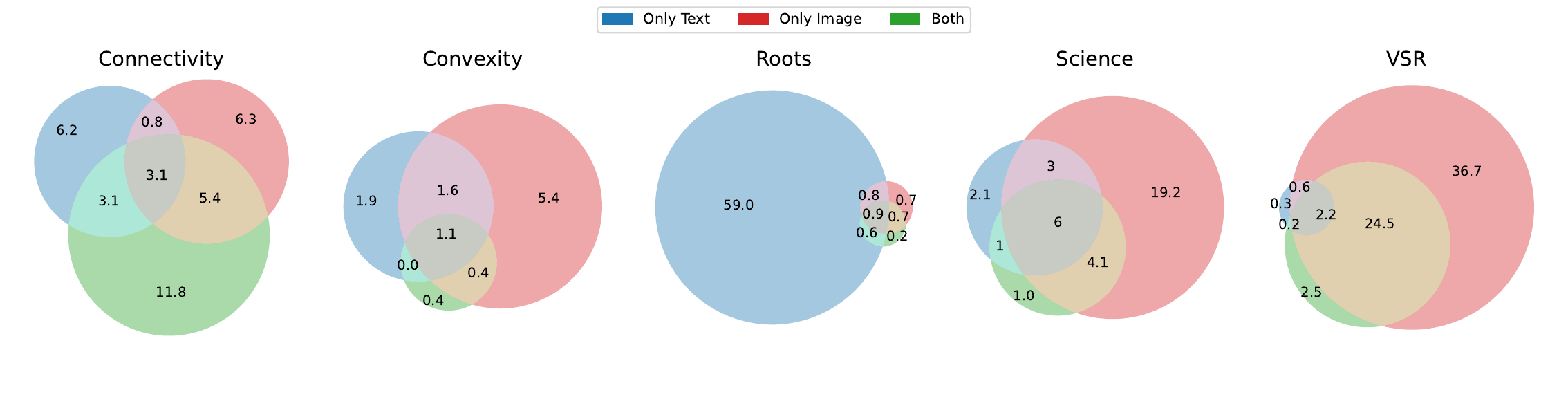}
    \vspace{-12mm}
    \caption{\textbf{Venn diagram of errors} displaying the percentage of mispredicted samples when GPT-4o mini is provided with only text, only image, and both text and image inputs.}
    \label{fig:gpt4o-mini-error}
\end{figure*}

\begin{figure*}[t!]
    \centering
    \includegraphics[width=\linewidth]{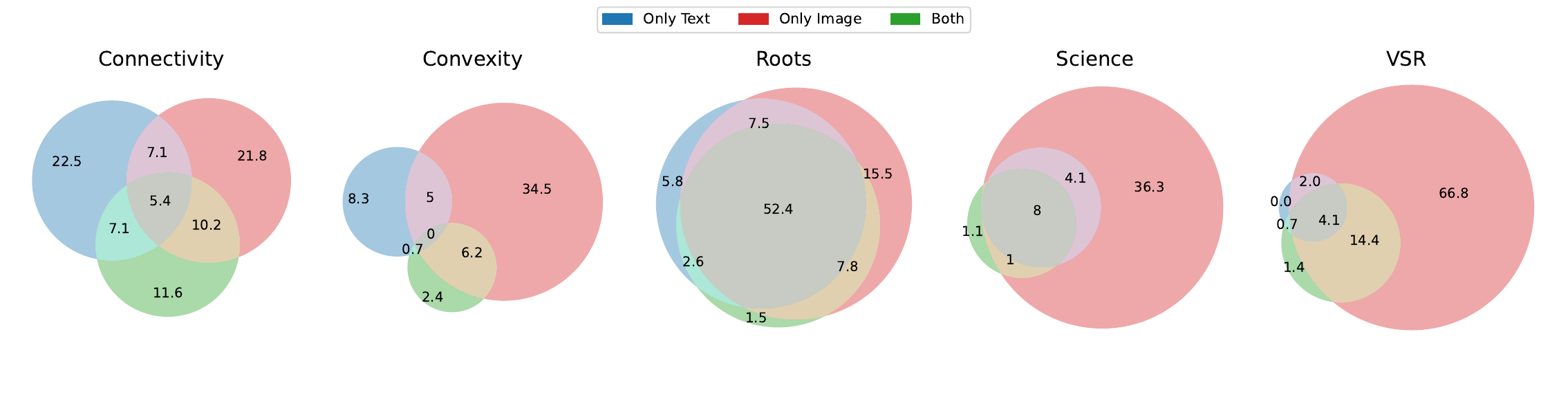}
    \vspace{-12mm}
    \caption{\textbf{Venn diagram of errors} displaying the percentage of mispredicted samples when Llama-3.2 90B is provided with only text, only image, and both text and image inputs.}
    \label{fig:llama-error}
\end{figure*}

\begin{figure*}[t!]
    \centering
    \includegraphics[width=\linewidth]{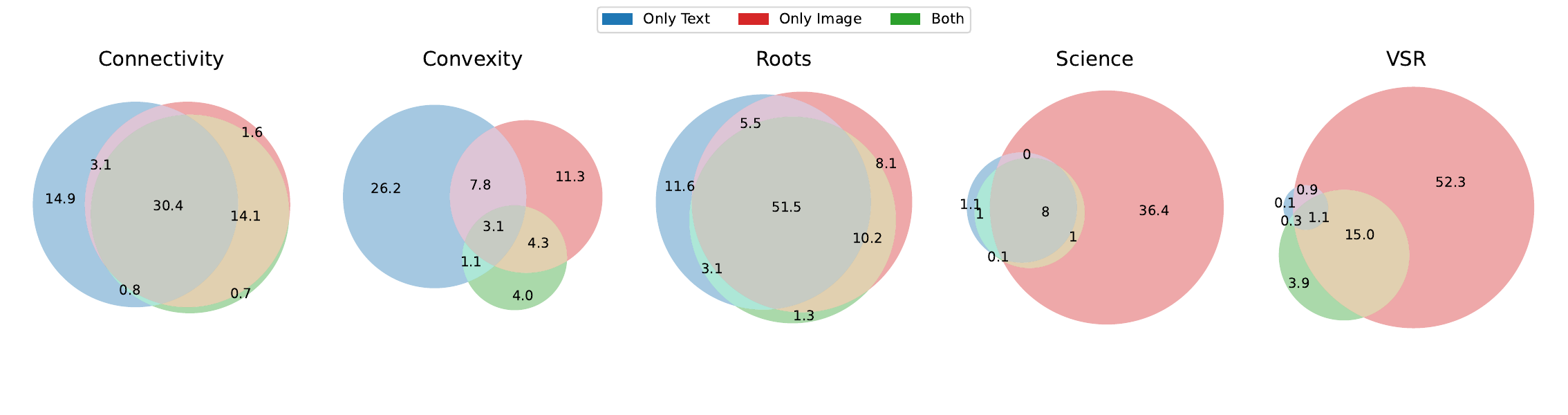}
    \vspace{-12mm}
    \caption{\textbf{Venn diagram of errors} displaying the percentage of mispredicted samples when Qwen2 72B is provided with only text, only image, and both text and image inputs.}
    \label{fig:q72-error}
\end{figure*}

\begin{figure*}[t!]
    \centering
    \includegraphics[width=\linewidth]{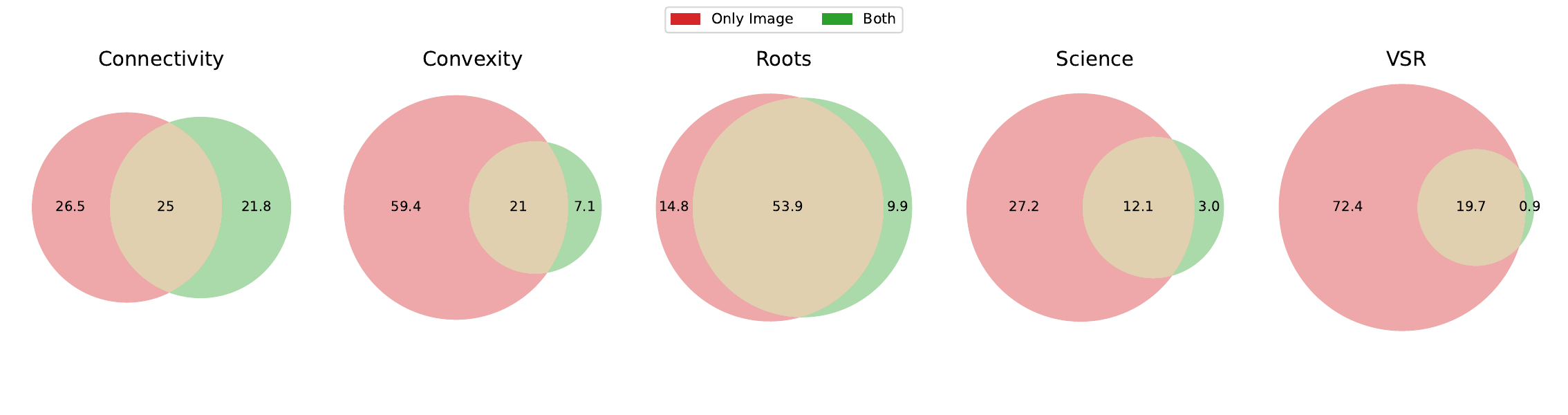}
    \vspace{-12mm}
    \caption{\textbf{Venn diagram of errors} displaying the percentage of mispredicted samples when Qwen2 7B is provided with only text, only image, and both text and image inputs.}
    \label{fig:q7-error}
\end{figure*}

We present a Venn diagram illustrating GPT-4o mini, Llama-3.2 90B’s, Qwen2 72B’s, Qand wen2 7B’s error percentages under three different input settings: text-only, image-only, and combined matching initial text-image inputs in Figures \ref{fig:gpt4o-mini-error}, \ref{fig:llama-error}, \ref{fig:q72-error}, and \ref{fig:q7-error}, respectively. We observe similar patterns to those seen with GPT-4o, particularly regarding the correlation between the models' unique and common error counts and their reported bias across different tasks.

\end{document}

%% file: math_commands.tex
\usepackage{amsmath,amsfonts,bm}

\def\eqref#1{equation~\ref{#1}}

\def\1{\bm{1}}

\DeclareMathAlphabet{\mathsfit}{\encodingdefault}{\sfdefault}{m}{sl}
\SetMathAlphabet{\mathsfit}{bold}{\encodingdefault}{\sfdefault}{bx}{n}

\usepackage{color}
\usepackage{times}
\usepackage{epsfig}
\usepackage{graphicx}
\usepackage{amsmath}
\usepackage{amsthm}
\usepackage{amssymb}
\usepackage{mathtools}
\usepackage{multirow}
\usepackage{bbm}
\usepackage{dsfont}

\usepackage[table, dvipsnames]{xcolor}

\usepackage[utf8]{inputenc}
\usepackage{booktabs,siunitx}
\usepackage{multirow}
\usepackage{multicol}
\usepackage{amsmath}
\usepackage{subcaption}
\usepackage{caption}
\usepackage{graphicx}
\usepackage{pifont}

\usepackage{tikz}
\usetikzlibrary{shapes,arrows,patterns}
\usetikzlibrary{positioning}
\usetikzlibrary{calc}

\newcommand{\cut}[1]{}

\newcommand{\postspace}{\vskip -3mm}
\newcommand{\minipostspace}{\vskip -2mm}

\definecolor{red}{RGB}{255, 117, 115}
\definecolor{green}{RGB}{171, 255, 175}